\documentclass[10pt,twocolumn]{article} 
\usepackage{simpleConference}
\usepackage{times}
\usepackage{graphicx}
\usepackage{amssymb}
\usepackage{url,hyperref}
\usepackage{algorithm}
\usepackage{algpseudocode}
\usepackage{amsmath}
\usepackage{graphics}
\usepackage{epsfig}

\begin{document}
\title{Pseudo-Representation Labeling Semi-Supervised Learning}
\author{Song-Bo Yang \\
\\
Taiwan Evolutionary Intelligence Laboratory\\
National Taiwan University \\
\\
r07921077@ntu.edu.tw  
\and
Tian-Li Yu \\
\\
Taiwan Evolutionary Intelligence Laboratory\\
National Taiwan University \\
\\
tianliyu@ntu.edu.tw  
}
\maketitle
\thispagestyle{empty}

% Abstraction======================================================================
\begin{abstract}
In recent years, semi-supervised learning (SSL) has shown tremendous success in leveraging unlabeled data to improve the performance of deep learning models, which significantly reduces the demand for large amounts of labeled data. Many SSL techniques have been proposed and have shown promising performance on famous datasets such as ImageNet and CIFAR-10. However, some exiting techniques (especially data augmentation based) are not suitable for industrial applications empirically. Therefore, this work proposes the pseudo-representation labeling, a simple and flexible framework that utilizes pseudo-labeling techniques to iteratively label a small amount of unlabeled data and use them as training data. In addition, our framework is integrated with self-supervised representation learning such that the classifier gains benefits from representation learning of both labeled and unlabeled data. This framework can be implemented without being limited at the specific model structure, but a general technique to improve the existing model. Compared with the existing approaches, the pseudo-representation labeling is more intuitive and can effectively solve practical problems in the real world. Empirically, it outperforms the current state-of-the-art semi-supervised learning methods in industrial types of classification problems such as the WM-811K wafer map and the MIT-BIH Arrhythmia dataset.
\end{abstract}

% Introduction=====================================================================
\begin{figure}[!t]
  \begin{center}
    \includegraphics[width=3.2in]{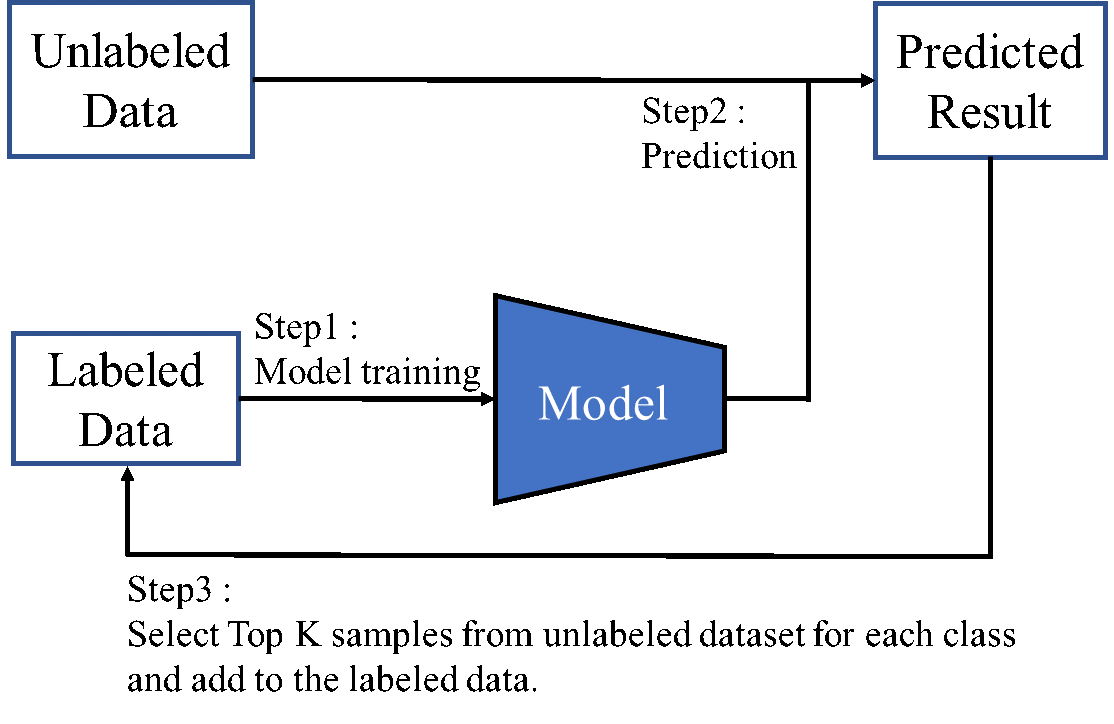}
  \end{center}
  \caption{\small A flow chart of pseudo batch labeling. It shows how to propagate small amount labels to unlabeled data iteratively in each class by pseudo-representation labeling.}
  \label{pseudo-labelling}
\end{figure}

\section{Introduction}
\label{Introduction}
Deep neural networks have achieved outstanding results in lots of computer vision challenge, such as object detection \cite{everingham2015pascal}, image classification \cite{deng2009imagenet}, object segmentation \cite{pont20172017}, and so on. However, these successes mentioned above are built on large amounts of label data, which is expensive to collect. Therefore, if a certain degree of knowledge can be extracted from a considerable quantity of unlabeled data, it must bring substantial commercial value. 

Semi-supervised learning \cite{chapelle2009semi} (SSL) explores unlabeled data to alleviate the problem of classifier overfitting caused by limited labeled data. 
In recent semi-supervised learning research, it is common to use consistent regularization on large amounts of unlabeled data to constrain model predictions to mitigate input noise \cite{berthelot2019mixmatch,xie2019unsupervised}. Due to the data outlier influence, this kind of method may not generate significant benefits on some real-world classification problems, e.g., wafer defect map classification and electrocardiography classification.
 \begin{figure}[!t]
  \begin{center}
    \includegraphics[width=3.2in]{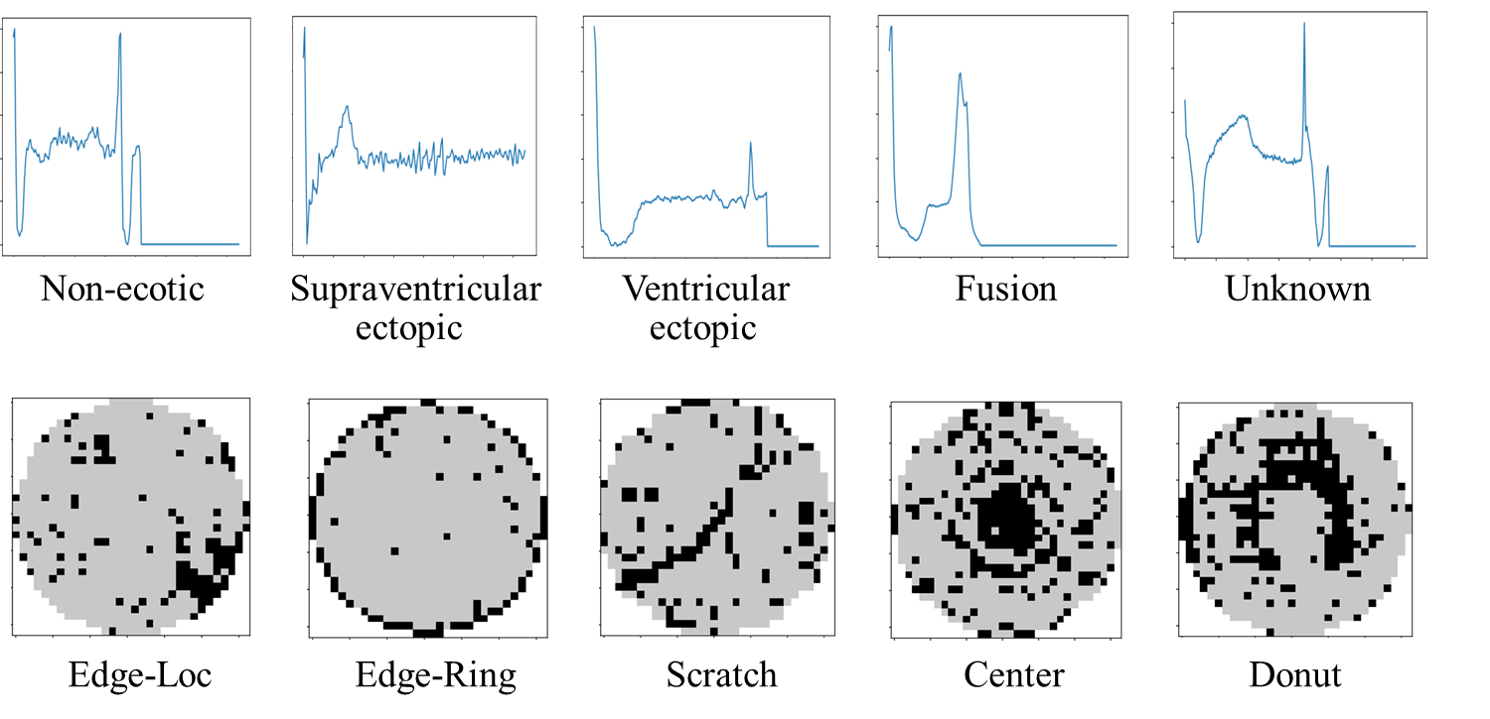}
  \end{center}
  \caption{\small Images of the WM-811K wafer map and the MIT-BIH Arrhythmia dataset. The WM-811K wafer map has seven classes, and the MIT-BIH Arrhythmia dataset has five classes.}
  \label{data}
\end{figure}
This work proposes a modified version of pseudo labeling \cite{lee2013pseudo}. In the beginning, the system trains a self-supervised model through unlabeled data to learn the representation of the data. It then combines supervised and unsupervised models and trains the classifier with both the labeled data and an increasing amount of unlabeled data with high confident pseudo labels. 
To ensure the correctness of the chosen pseudo labeled data at each iteration, only a small batch of pseudo labeled data, which has the highest probability at a time to add to the supervised part.
This framework is competitive to the novel semi-supervised learning methods, and show outstanding results on industrial datasets such as the WM-811K wafer map and the MIT-BIH Arrhythmia dataset. We demonstrate these two datasets in Figure \ref{data}.

A fundamental assumption in semi-supervised learning is that labeled data and unlabeled data belong to the same distribution. However, the actual situation in real-world problems may not be like this, but there are many unlabeled samples beyond distribution. 
Pseudo-representation labeling only adds a small amount of unlabeled data with high classification confidence, instead of all of it. In addition to the problem encountered above, this paper also investigates how many unlabeled data should be added each time to get the maximum efficiency. Bear in mind that every time we add unlabeled data into the training set, a certain degree of noise is generated. As iteration continues, the noise gradually increases, which causes the performance of the model to decrease.

Our main contributions can be summarized as follows:
\begin{description}
  \item[$\bullet$]\hspace{0.3cm}This paper proposes a new integrated architecture to unify representation learning and semi-supervised learning approaches.
  \item[$\bullet$]\hspace{0.3cm}This paper shows that it is not suitable to add all unlabeled data at once. Selecting a small amount of unlabeled data with high confidence in each class iteratively can improve the overall model performance. 
  \item[$\bullet$]\hspace{0.3cm}Pseudo-representation labeling outperforms the existing semi-supervised learning techniques in the WM-811K wafer map and the MIT-BIH Arrhythmia dataset.
\end{description}

The remainder of this paper is organized as follows. Section~\ref{Related Work} reviews the recent semi-supervised learning and self-supervised learning researches. 
Section \ref{Our Method} introduces the framework and the algorithm of pseudo-representation labeling. More details and experimental results are evaluated in Section \ref{Experiment}. 
Finally, Section \ref{Conclusion} summarizes the entire work.

% Related Work========================================================================
\section{Related Work}
\label{Related Work}
Pseudo-representation labeling is built on recent semi-supervised learning and self-supervised learning approaches. To make the overall background more clear, this section reviews the current state-of-the-art in both fields.
\subsection{Semi-Supervised Learning}
Semi-supervised learning is a technique leveraging unlabeled data to improve the performance of deep learning models. A common assumption is that unlabeled data comes from the same distribution as labeled data. Survey according to \cite{berthelot2019mixmatch}, recent semi-supervised learning approaches can be mainly divided into consistency regularization, entropy minimization, and traditional regularization. Next, this section explains these categories one by one.

\subsubsection{Consistency Regularization}
~~~The main idea of consistency regularization is that, after the input to be undergone data augmentation, its class semantic should remain the same. Due to the perturbations of unlabeled data should not influence class consistency. This idea is widely utilized in many state-of-the-art semi-supervised algorithms such as UDA \cite{xie2019unsupervised}, Mixmatch \cite{berthelot2019mixmatch}, and EnAET \cite{wang2019enaet}. 
However, to overcome the noise caused by augmentation and maintain the stability of the model, it is usually essential to use a relatively stable loss term for the consistency part.
For example, VAT \cite{miyato2018virtual} adds KL divergence in the unsupervised loss term
\begin{equation}
    L = P(y|Augment(x))\log{\frac{P(y|Augment(x))}{P(y|Augment(x))'}}
\end{equation}
and Mixmatch adds L2 norm in the unsupervised loss term.
\begin{equation}
    L = \|P(y|Augment(x)) - P(y|Augment(x))'\|_2^2
\end{equation}
Different from the previous methods of adding loss term to unlabeled data, “Mean Teacher” \cite{tarvainen2017mean} using two models to update the weights with exponential moving average (EMA), and force the prediction results of both models to be the same. This design is useful in oversoming the overfitting problem on a single model. In addition to loss term related methods, data augmentation is also an essential technique for consistency regularization. Proposed in UDA \cite{xie2019unsupervised},  RandAugment \cite{cubuk2019randaugment} is a data augmentation method with two hyperparameters, which control the intensity and amount of augmentation respectively. 
EnAET \cite{wang2019enaet} proposed an ensemble method to combine spatial and non-spatial transformations to implement augmentation.

\subsubsection{Entropy Minimization}
~~~Central idea through entropy minimization believes that unlabeled data, which is located in high confidence regions, belong to nearby labels, and the decision boundary locates near low-density areas. The most classic research of them is the pseudo labeling \cite{lee2013pseudo}. Pseudo labeling uses the predicted label as the real label through cross-entropy 
\begin{equation}
    -P(y|x)\log{P(y|x)}
\end{equation}
and gradually increase the proportion of its loss during the training process. 
However, an essential point in this method is that if the error rate of the model at the beginning is too high, the subsequent training process only gets worse. Because our assumption is usually a tiny amount of labeled data and a lot of unlabeled data, this situation happens often.
Pseudo-representation labeling borrows and improves the idea of pseudo labeling. The detailed explanation is be placed in Section \ref{Experiment}. 

\subsubsection{Traditional Regularization}
~~~Traditional regularization refers to directly restricting the loss term of the model to make the prediction curve smoother, and this technique generalizes the model to avoid the overfitting caused by a small amount of data. For example, weight decay \cite{krogh1992simple} is a kind of regularization widely used in machine learning models. 
\begin{equation}
    L(\theta) = L(\theta)' + \frac{1}{2}\lambda\mathop{\sum_{i=1}^{n}}\theta_i^2
\end{equation} 
Many new techniques have also been discussed in recent years, such as \cite{nakamura2019adaptive,loshchilov2017decoupled,loshchilov2018fixing}.
Traditional regularization is simple and easy to implement, and thus it becomes the most commonly used technique among the three kinds of semi-supervised methods. However, traditional regularization can not learn the information from unlabeled data, but it only needs to adjust the loss term of the model. In our thought, the quantity of information added is insufficient.

\subsection{Self-Supervised Learning}
Self-supervised learning is a simple but powerful technique able to extract information from unlabeled data. We can design appropriate problems for unlabeled data in many different ways according to different scenarios as long as one can define the characteristics of data. In recent years, self-supervised learning has become popular due to an excellent pre-trained model without the artificial image label is valuable. This is not only helpful for unsupervised learning but also be advantageous to semi-supervised learning.

However, designing an appropriate self-supervised problem is not an easy job because it must consider the characteristics of the training data. For example, image rotation \cite{gidaris2018unsupervised} is a well-known self-supervised learning task, which rotates the image through different angles, and uses the rotation angle as the label. It can achieve good results on most types of data. However, if the semantic meaning of training data is not significantly altered after rotation (for example, the data is round), this technique does not work well. 
Predicting the relative position of image patches is also a self-supervised method \cite{doersch2015unsupervised}, by cutting the picture into grids and randomly sampling two patches to determine the relative position of each other. This method performs much better on image detection than image classification, and once again shows that it is critical to design appropriate self-supervised tasks based on different scenarios. Compared with image classification, relative position prediction is more reasonable on object detection. In addition, there are also methods for converting grayscale images to color \cite{zhang2016colorful} or predicting the transformation of images \cite{zhang2019aet} to achieve image representation learning.

All of the above methods have shown that self-supervised learning can significantly improve model performance. In practice, one can fix the pre-train model weight to do the classification or just use labeled data to fine-tune the pre-train model. This paper explains how to use self-supervised learning to integrate pseudo-representation labeling in Section \ref{Our Method}.

% our method============================================================
\section{Pseudo-Representation Labeling}
\label{Our Method}

\begin{figure}[!b]
  \begin{center}
    \includegraphics[width=3.5in]{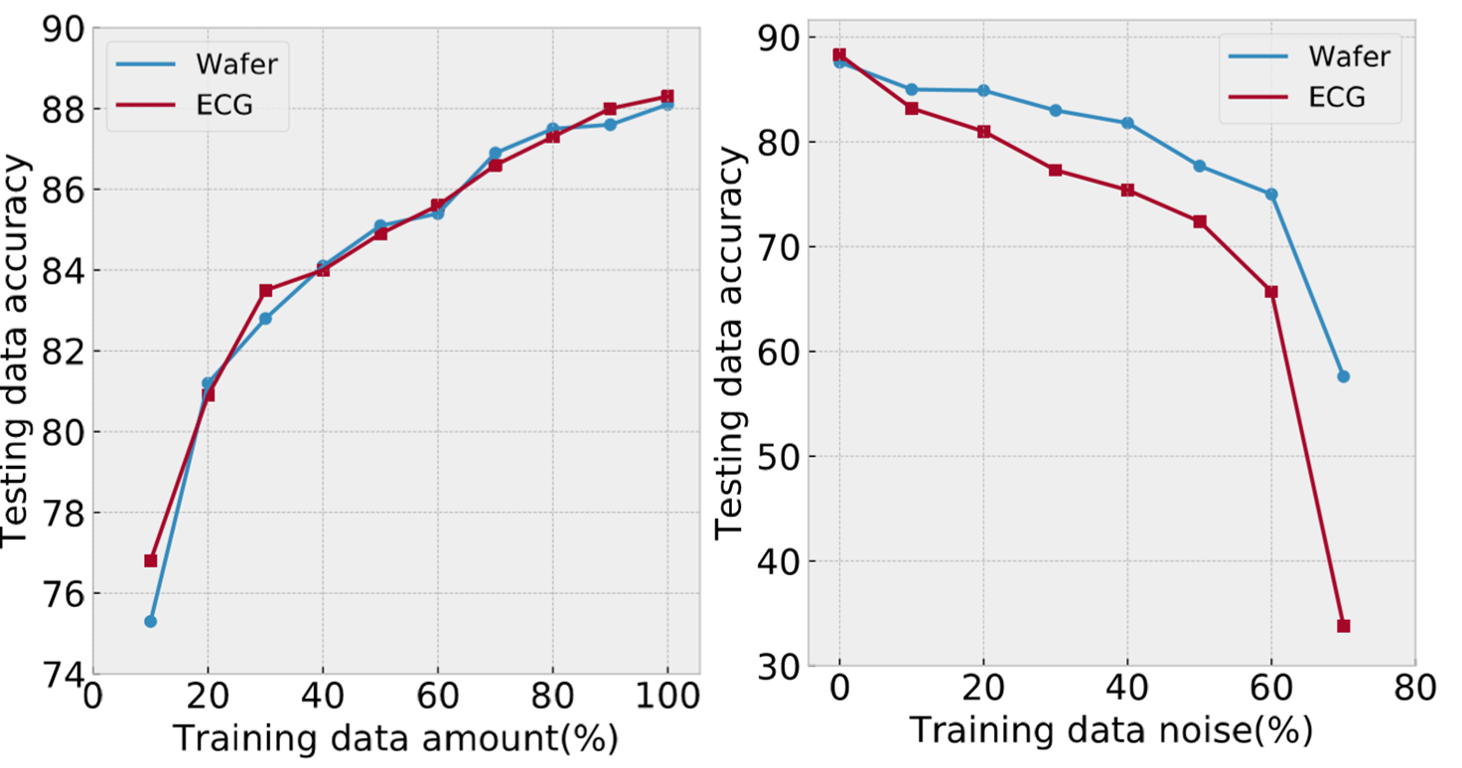}
  \end{center}
  \caption{\small Trend charts of data amount and data noise. 
  We use WM-811k wafer map and MIT-BIH Arrhythmia heartbeat graph for image classification, and these experiments are implemented with data augmentation. 
  The left figure shows the amount of data and accuracy increase at the same time. 
  The right figure shows that noise reduces accuracy. 
  In the process of adding data, the two kinds of mechanisms conflict with each other.}
  \label{increase}
\end{figure}

This section introduces the technique of pseudo-representation labeling, a flexible and straightforward framework that combines self-supervised learning and pseudo labeling techniques to an integrated architecture. Our framework applies to semi-supervised learning in image classification, and it can be mainly divided into two parts. The first part is to gradually spread from a small amount of label data to unlabeled data, and the second part is to combine the self-supervised learning technique to improve the overall performance.    

\subsection{Pseudo-batch labeling}
In machine learning tasks, as the number of labeled data increases, the classifier accuracy is also increased due to the benefit from labeled data. On the contrary, when the noise in the label data enlarges, the classifier accuracy decreases. We use WM-811k and MIT-BIH Arrhythmia dataset to conduct our experiment and show these two kinds of mechanisms in Figure \ref{increase}. These phenomenons are simple and intuitive, which are also the problem to be encountered in the real world. 
When pseudo labels are assigned to unlabeled data and merged with labeled data, the new labeled data contain a certain degree of incorrect labels that form the labeled data noise. The noise causes a decrease in accuracy and conflict with the benefit of increased label data. 
It shows that if the accuracy decline cause from data noise is greater than the benefit brought by adding unlabeled data, this problem may not be in line with our scenario.
\begin{figure}[!t]
  \begin{center}
    \includegraphics[width=2.4in]{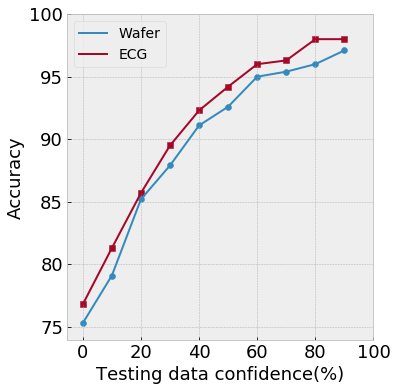}
  \end{center}
  \caption{\small A graph shows the impact of confidence on the accuracy. We use WM-811k wafer map and MIT-BIH Arrhythmia dataset for image classification, and these experiments are implemented with data augmentation. The results show that the testing data selected through higher prediction confidence by this classifier indeed have higher accuracy.}
  \label{confidence}
\end{figure}
Therefore, what we need to do is to suppress the noise of the data as much as possible during the process of adding labeled data. 

It was mentioned in pseudo labeling that data selected by prediction confidence is an excellent metric to minimize the prediction entropy and to reduce the input noise. Our experiments measure how different prediction confidences in the testing data affect accuracy. In Figure \ref{confidence}, wafer defect maps and ECG heartbeat graph are used for image classification training. 
The result shows that selected testing data through high prediction confidence can indeed find data with smaller noise and higher accuracy. 
So the question this section discuss next becomes how much unlabeled data should be added each iteration to maximize the performance of our classifier. 
If too much unlabeled data be added per round, the noise rises too fast, and the model performance decrease. On the other hand, if too little be added each time, it leads to inefficiency and high training cost. We show and discuss the results of adding unlabeled data at different sizes each iteration in Section \ref{Experiment}.

\begin{figure}[!b]
  \begin{center}
    \includegraphics[width=3.2in]{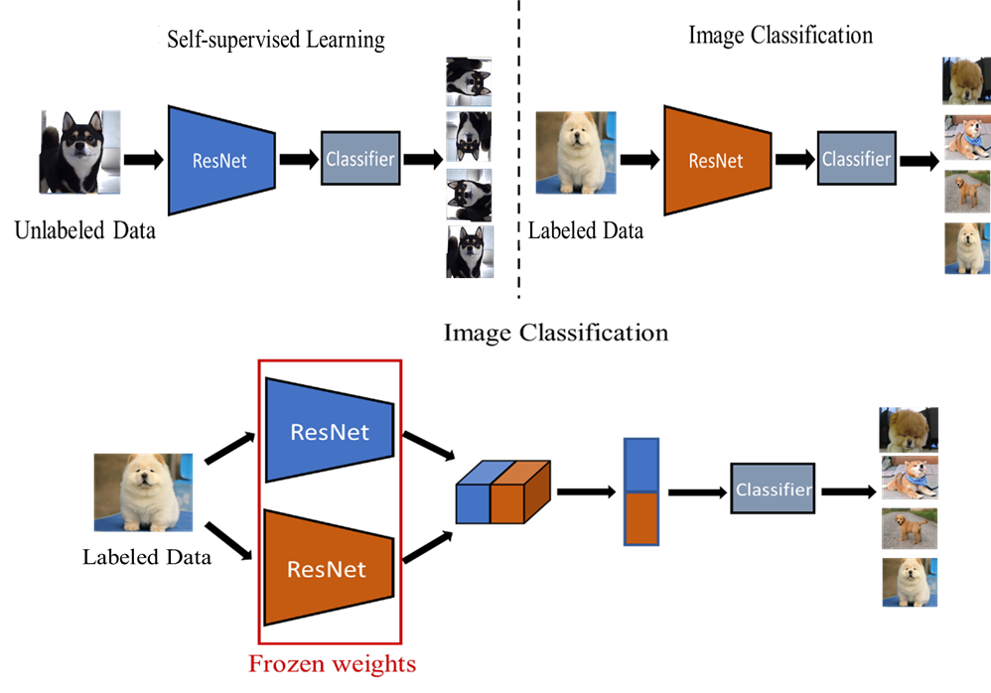}
  \end{center}
  \caption{\small A framework to combine self-supervised learning and semi-supervised learning. This framework can be divided into two parts. First, it trains a classifier on unlabeled data with self-supervised learning. Next, it trains labeled data with supervised learning. By fixing the parameters of these two nets and concatenating them, this framework generates a classifier to do image classification.}
  \label{framework}
\end{figure}

\begin{algorithm}[!t]
  \caption{Pseudo-representation labeling}
  \label{alg1}
  \begin{algorithmic}[1]
    \Require
      labeled data pair $\chi = (L_x,L_y)$, unlabeled data $U_x$, number of augmentation $k$, increasing ratio $\alpha$, labeled data amount N.
    \State $\chi' = \emptyset$  
    \For{$i=1$ to $k$}
        \State $\chi' = (\chi')~\cup$ Augmentation$(\chi)$
    \EndFor
    \State $M_u\leftarrow$ train a representation classifier from $U_x$ 
    \While {true}
        \State $M_l\leftarrow$ train a classification classifier from $\chi'$
        \State $E_l \leftarrow$ extraction $\chi'$ feature embedding from $M_l$
        \State $E_u \leftarrow$ extraction $\chi'$ feature embedding from $M_u$
        \State $W$ = Concatenate($E_l,E_u$)
        \State $M_w\leftarrow$ train a classification classifier from $W$ 
        \State $U_s$ = select $\alpha \times N$ unlabeled samples from $M_w(U_x)$ 
        
        \hspace{0.6cm}for each class
        \State $U_x$ = $U_x-U_s $ 
        \State $\chi'$ = $\chi'~\cup (U_s,M_w(y | U_s))$ 
    \EndWhile
  \end{algorithmic}
\end{algorithm}

\subsection{Representation Learning Integration}

Our system integrates the information learned from unlabeled data during the training process to improve our classifier performance. To achieve this, we designed a framework that strengthens the process of label selection in combination with self-representation learning. This framework is shown in Figure \ref{framework}, and it can be divided into two parts. First, it trains a classifier on unlabeled data with self-supervised learning. Next, it trains labeled data with supervised learning. By fixing the parameters of these two nets and concatenating them, this framework generates a classifier to do image classification. In our framework, the appropriate self-representation learning method is chosen according to different types of data. 

The algorithm is as fallows Algorithm \ref{alg1}. First, $N$ times of augmentation on labeled data $\chi$ and unlabeled data $U_x$ are performed respectively. Then, we train a self-supervised model on unlabeled data and a supervised model on labeled data according to the image classification task. Feature embedding vectors are extracted based on these two models, and the feature embedding vectors are used to train the image classification classifier. Finally, $\alpha \times N$ unlabeled data is selected in each class to be added to the labeled data according to the prediction confidence by this classifier.

\subsection{Feature Space Data Augmentation}
\label{FSDA}
 \begin{figure}[!t]
  \begin{center}
    \includegraphics[width=3.3in]{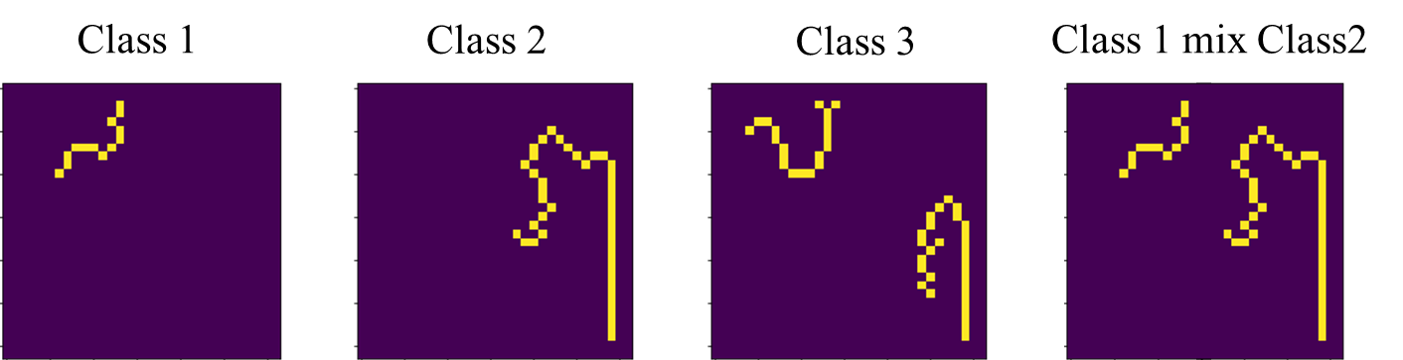}
  \end{center}
  \caption{\small An example shows Mixup may not work well in some industrial type of dataset. It is a classification problem to distinguish the crack is on the right side, left side, or both. If the data in class 1 and class 2 are combined by mixup, and the one-hot label should not be the ratio of these two but class 3. }
  \label{phone}
\end{figure}
Mixup \cite{zhang2017mixup} is a milestone data augmentation technology to increase the training data amount. It is useful and easy to implement and widely used in recent semi-supervised learning algorithms. However, Mixup may not work well under specific scenarios, and the example is shown in Figure \ref{phone}. Figure \ref{phone} shows a synthetic crack dataset. Class 1 and Class 2 represent the crack on the left and right, respectively, and Class 3 means both sides. If Class 1 and Class 2 are combined by Mixup, the one-hot label should not be the weighted value of these two but Class 3. This type of challenge appears on datasets for industrial applications frequently, and it is necessary to find the solution to solve this problem.

In \cite{devries2017dataset}, three different data augmentation methods are performed in the feature space, but the improvement is not apparent. Variational auto-encoder (VAE)  \cite{kingma2013auto} may be a solution to make feature space data augmentation more reasonable. By adding noise to the embedding layer to train the model, the transition between instances of different two classes is slow, and the mixed image contains the characteristics of both classes. This property makes Mixup in the feature space more applied to the industrial types of datasets.
Thus, we adopt Mixup by VAE to perform data augmentation at the embedding layer instead of the input layer in the framework.
% Experiment====================================================================
\section{Experiments}
\label{Experiment}
In this section, we explain the relevant experimental results.
Section \ref{details} introduces the experiment implementation details. Section \ref{noise} compares the impact of noise according to different amounts of labeled data. Section \ref{alpha} records the influence of adding unlabeled data under different batch sizes. The importance of the self-supervised part is revealed in Section \ref{self-supervised}. Section \ref{mixlayer} implements feature space Mixup in different model layers. Finally, Section \ref{result} compares the experimental results with other state-of-the-art semi-supervised learning algorithms. 

\begin{figure}[!b]
  \begin{center}
    \includegraphics[width=3.2in]{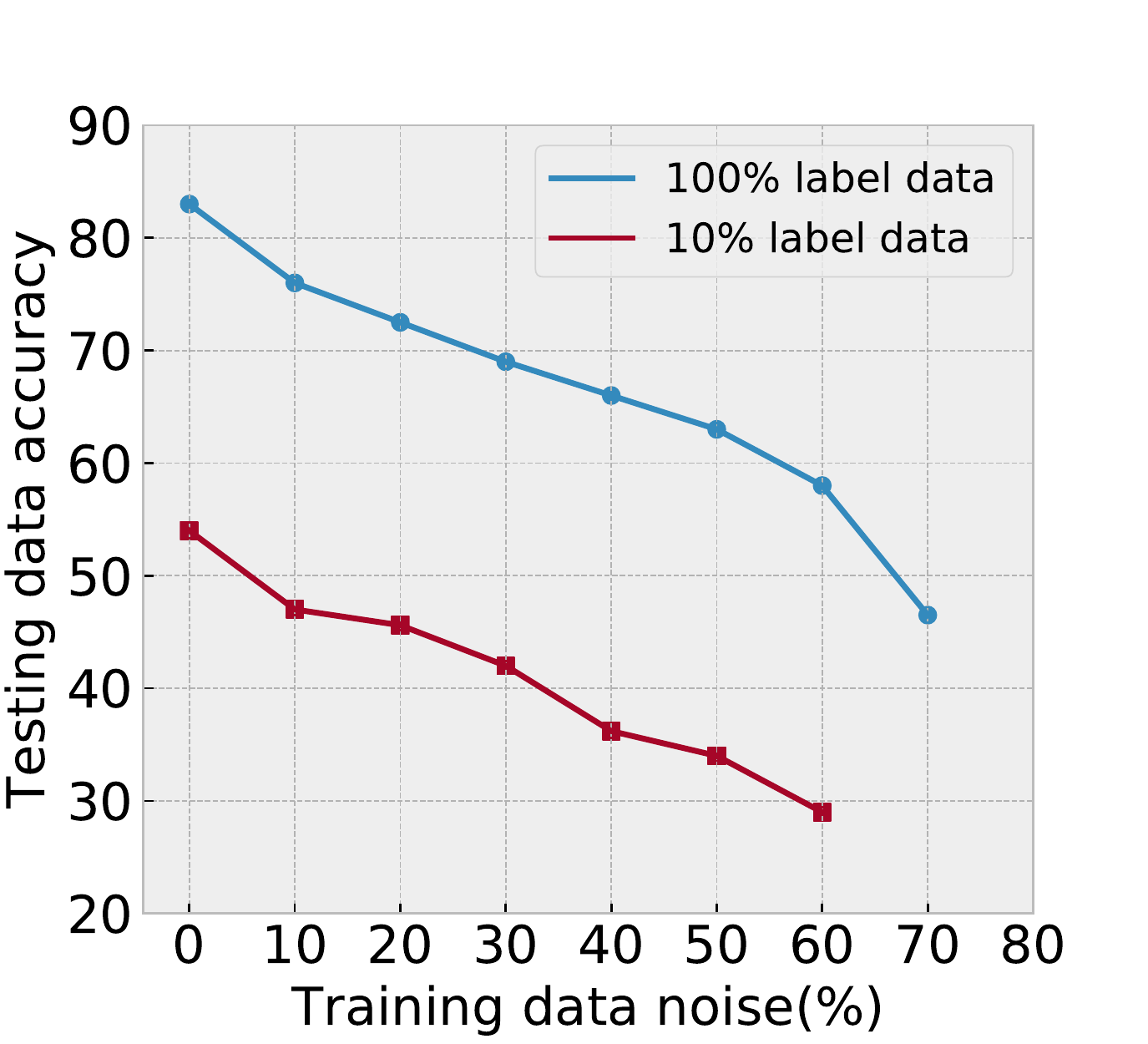}
  \end{center}
  \caption{\small  The tendency of accuracy on different levels of noise. It shows that the two descending curves are very similar under different percentages of data.}
  \label{evaluate}
\end{figure}

\begin{figure}[!t]
  \begin{center}
    \includegraphics[width=3.2in]{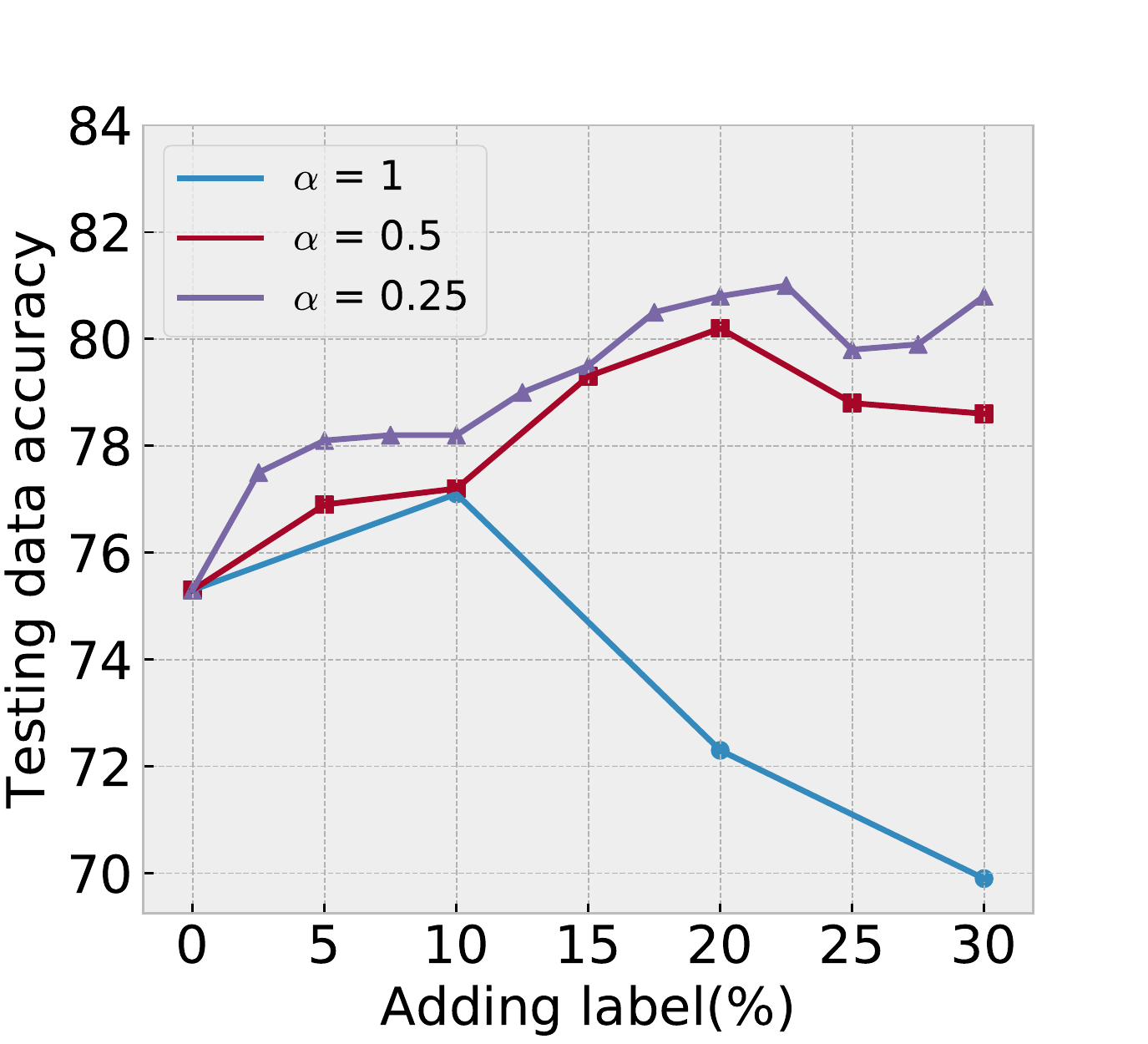}
  \end{center}
  \caption{\small Pseudo-representation labeling accuracy under different batch size $\alpha$. We use 315 labeled wafer defect maps and 2835 unlabeled wafer defect maps to implement iterative pseudo labeling semi-supervised classification. When the $\alpha$ is too large ($\alpha = 1$), the accuracy initially increases slightly, and then the accuracy decreases due to the noise. When the alpha is small ($\alpha<=0.5$), the growth of accuracy is more stable. After growing to a certain level, the accuracy still declines due to the increased noise.}
  \label{compare}
\end{figure}

\subsection{Implementation Details}
\label{details}
{\bf Datasets}: In our experiment, the WM-811K wafer map and the MIT-BIH Arrhythmia dataset from Kaggle are used as our assessed benchmarks. For WM-811K, due to the uneven class quantity and image size, we randomly sample 3150 images from seven classes and resize them into 32x32. For ECG, the MIT-BIH arrhythmia dataset is a five classes classification problem, and the number of sample data is 2000. Both of the unlabeled data are randomly sampled from the initial dataset, and the rest are treated as labeled data. Apart from this, the experiments also independently sample the testing data.

{\bf Experiment details}:
Wide ResNet-28 \cite{zagoruyko2016wide} model is used to implement these experiments. For the hyperparameters, these experiments set the dropout rate as 0.2, the weight decay as 0.0005, and the batch size as 64 at each update. We use an initial learning rate of 0.1 and Adam as an optimizer to update the model weights. For the augmentation strategy, we apply horizontal flipping, vertical flipping, and rotation in WM-811K, and only horizontal flipping, vertical flipping in ECG. For both datasets, autoencoder is adopted to self-learned data representation. 
\begin{figure*}[!t]
  \includegraphics[width=7.05in]{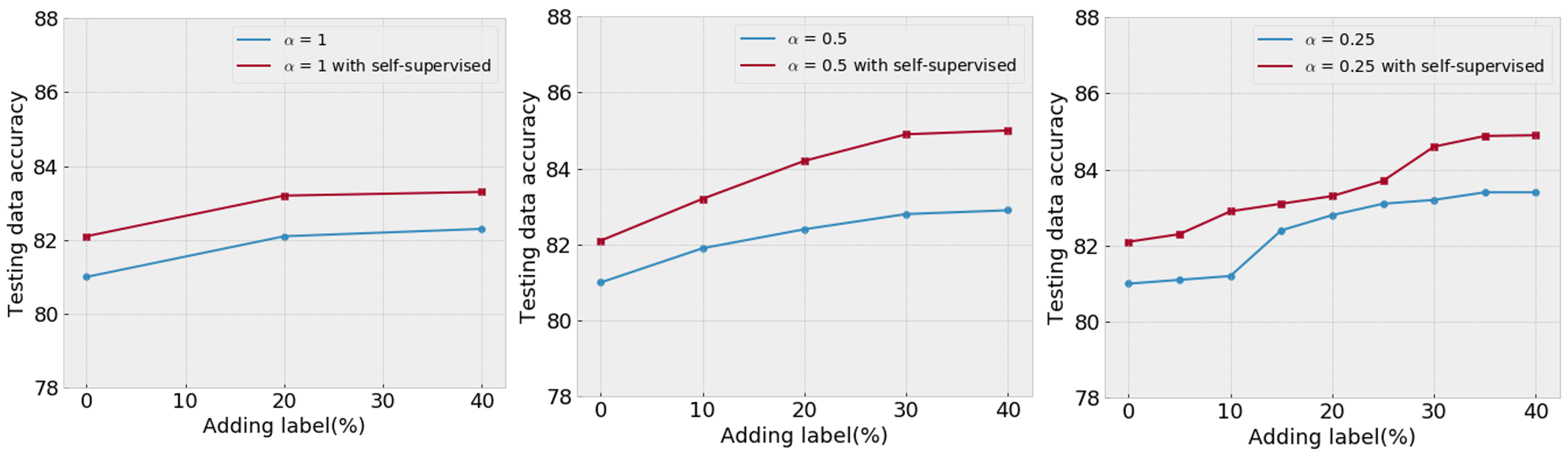}
  \begin{center}
  \caption{\small  Three charts comparing the improvement with different added unlabeled data sizes. We used 630 labeled wafer defect maps and the rest as unlabeled data. Experiments show that adding the representation of unsupervised data during training iteration helps model performance.}
  \label{self}
  \end{center}
\end{figure*}

\subsection{Noise Evaluation}
\label{noise}
In the Section \ref{Our Method}, it is mentioned that noise is generated during the process of adding unlabeled data to the labeled data, and this phenomenon conflict with the benefit of increased label data. However, in reality, the user does not know how much noise affects their data in advance, because they only hold a small part of the data on hand. In Figure \ref{evaluate}, these experiments take only 10\% of the WM-811K data and add different levels of noise to observe the tendency of accuracy. Unexpectedly, the 10\% data declining accuracy is quite similar to the declining accuracy calculated from 100\% data. According to the results, the user can try to estimate how much the noise influences the accuracy by using a small amount of existing data when we do not know the distribution of unknown data.
\begin{figure}[!t]
  \begin{center}
    \includegraphics[width=3.2in]{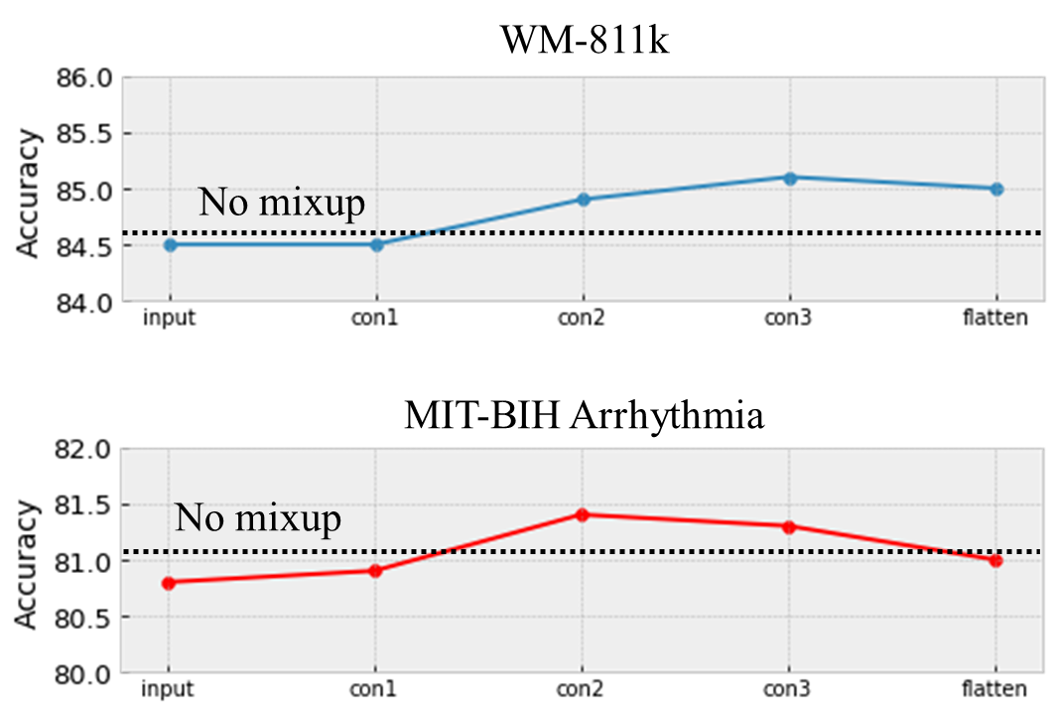}
  \end{center}
  \caption{\small Mixup according to different model layers to compare the training performance. In the results, compared to doing mixup in the input layer, feature space Mixup outperforms in both datasets.}
  \label{FeatureSpace}
\end{figure}
\subsection{Utilized Unlabeled Data Size}
\label{alpha}
Now we investigate the impact of different sizes of labeled data added per iteration. It was mentioned earlier that noise harms the benefits of labeled data. Here we use experiments in Figure \ref{compare} to illustrate how it happened. 
In the experiment shown in Figure \ref{compare}, 315 labeled wafer defect maps and 2835 unlabeled wafer defect maps are used to implement semi-supervised classification. This experiment has not yet added self-supervised technology, but only selects data based on prediction confidence.

The accuracy of the initial model trained with the labeled data is only 75.3\%, and we add unlabeled data according to different sizes. Given symbol $\alpha$ is the ratio that controls the size, and $N$ is the labeled data amount. In each round, $\alpha \times N$ unlabeled data is selected in each class to be labeled data. When the $\alpha$ is large ($\alpha = 1$), the accuracy increases slightly when trained with some pseudo-label data, and then decreases with more pseudo-label data due to noise. When the alpha is small ($\alpha<=0.5$), the growth of accuracy is more stable. After growing to a certain level, the accuracy still declines due to the increased noise. To summarize that small unlabeled data size can make the model grow more steadily, and we should use the validation set to select a batter model before the accuracy drops.

\begin{table}[!b]
\centering  
\begin{tabular}{lccc}  
\hline
Method & & \hspace{-3cm}Error Rates\\
\hline
 &Wafer (630 Label) &ECG (200 Label) \\ \hline  
Supervised &19.5 &23.2 \\         
MixMatch \cite{berthelot2019mixmatch} &18.8 &20.4 \\         
EnAET \cite{wang2019enaet} &16.6 &22.6 \\        
Curriculum Labeling \cite{cascante2020curriculum} &17.9 &21.7\\ \hline
Pseudo-Representation &\bf{15.1} &\bf{18.8}\\ \hline
\end{tabular}
\caption{\small We use the WM-811K wafer map and the MIT-BIH Arrhythmia dataset to evaluate the error rate of the current semi-supervised state-of-the-art methods.}
\label{tt}
\end{table}

\subsection{Self-Supervised Learning Improvement}
\label{self-supervised}
To improve this framework, in this step, self-supervised technology is joined to our framework, and the improvement is compared with different added unlabeled data sizes. We used 630 labeled wafer defect maps and the rest as unlabeled data. Besides, auto-encoder is chosen to be the self-supervised task in Figure \ref{self}. Experiments show that adding the representation of unsupervised data during training iteration helps model performance and accuracy achieve a certain degree of growth in all different added unlabeled sizes. 

\subsection{Mixup in Different Model Layer}
\label{mixlayer}
In Section \ref{FSDA}, we propose to adopt feature space Mixup to perform data augmentation at the embedding layer instead of the input layer. Thus, another question is which layer to execute on has higher performance. We perform the experiment in the input layer, convolution 1, convolution 2, convolution 3, and flatten layer, respectively, and the results are shown in Figure \ref{FeatureSpace}. The results show that Mixup in the input layer may not as well as a result without Mixup in these two datasets, and the best results occur in the second and third convolution layer, respectively.
\cite{zhang2017mixup} recommends that Mixup in the input layer or the earlier convolution layer is a good choice. Based on the results of our experiments and the experiments in \cite{zhang2017mixup}, we suggest performing Mixup in the second convolution layer as default to implement feature space Mixup.

\subsection{Results Comparison}
\label{result}
Finally, the WM-811K wafer map and the MIT-BIH Arrhythmia dataset are used to evaluate pseudo-representation labeling. Implementing semi-supervised learning with 630 labels and 200 labels respectively, we measure the error rate of the current semi-supervised state-of-the-art methods in the Table \ref{tt}. The results show that the pseudo-representation labeling outperforms the exiting semi-supervised learning techniques in these two datasets. 

% Conclusion
\section{Conclusion}
\label{Conclusion}
This paper proposed pseudo-representation labeling, a flexible SSL framework that combines self-representation learning with pseudo labeling to improve the performances of the image classification models. The system also adopted Mixup to perform data augmentation at the embedding layer instead of the input layer. Empirically, pseudo-representation labeling is more intuitive and effective than existing approaches for industrial applications, such as the WM-811K wafer map and the MIT-BIH Arrhythmia dataset. As for future work, we aim to ensemble different kinds of self-supervised methods to enhance the representations. Also, we would like to investigate different augmentation methods for pseudo-representation labeling.

\bibliographystyle{abbrv}
\bibliography{refs}
\end{document}